\documentclass[journal]{IEEEtai}

\usepackage[colorlinks,urlcolor=blue,linkcolor=blue,citecolor=blue]{hyperref}

\usepackage{amsmath,amsfonts}
\usepackage{algorithmic}
\usepackage{algorithm}
\usepackage{array}
\usepackage[caption=false]{subfig}
\usepackage{textcomp}
\usepackage{stfloats}
\usepackage{url}
\usepackage{verbatim}
\usepackage{graphicx}
\usepackage{cite}
\usepackage{tabularx}
\usepackage{booktabs}
\usepackage{multirow}
\usepackage{makecell}
\usepackage[table]{xcolor}
\usepackage{bbding}
\usepackage{balance}
\DeclareMathOperator{\bigcdot}{\cdot}
\usepackage{pifont}

\begin{document}

\title{Efficient Cross-Scale Invertible Hiding Network with Spatial-Frequency Collaboration and Non-Invertible Mechanism} 

\author{Junxue Yang, Xin Liao$^{*}$
	\thanks{This work is supported by National Key R\&D Program of China (Grant Nos. 2024YFF0618800, 2022YFB3103500). National Natural Science Foundation of China (Grant No. U22A2030). Hunan Provincial Funds for Distinguished Young Scholars (Grant No. 2024JJ2025), Hunan Provincial Key Science and Technology Program (Grant No.2025QK2008), Hunan Provincial Key Research and Development Program (Grant Nos. 2024AQ2027, 2025AQ2022).}
	\thanks{Junxue Yang, Xin Liao are with the College of Computer Science and Electronic Engineering, Hunan University, Changsha, China (e-mail: \{junxueyang, xinliao\}@hnu.edu.cn).} 
	\thanks{$^{*}$Corresponding authors: Xin Liao}}


\markboth{Journal of IEEE Transactions on Artificial Intelligence, Vol. 00, No. 0, Month 2026}
{Efficient Cross-Scale Invertible Hiding Network with Spatial-Frequency Collaboration and Non-Invertible Mechanism}

\maketitle

\begin{abstract}
Recently, deep learning-based steganographic models have demonstrated promising potential for hiding a full-size secret image within a cover image of the same size, a technique also known as image hiding. Invertible neural networks (INN)-based image hiding has emerged as an important branch. It treats concealing and revealing as a pair of inverse problems on image domain transformation and uses INN's forward and backward processes to address them. Due to architectural constraints, existing INN-based methods suffer from single-scale and single-domain feature extraction and limited nonlinear representation capability, resulting in inferior image quality. To mitigate these limitations, we propose an efficient cross-scale invertible hiding network with the spatial-frequency collaboration and the non-invertible mechanism, termed CrosInv. CrosInv exploits cross-scale and spatial-frequency collaborative features while enhancing nonlinear representation. Specifically, we introduce a cross-scale invertible module that bijectively maps inputs to cross-scale representations. To effectively integrate spatial and frequency information, the cross-scale invertible module employs pixel shuffle, Haar wavelet transformation, and their inverse operations for scale transformation. Furthermore, the non-invertible cross dense module is combined to enhance the nonlinearity. Comprehensive experiments verify the effectiveness and superiority of the proposed CrosInv.
\end{abstract}

\begin{IEEEImpStatement}
Images are ubiquitous on the Internet, and many of them contain private content ranging from personal information to business and government data. These images are vulnerable to interception during online transmission. Image hiding can protect private content by embedding a full-size secret image within an ordinary-looking cover image without arousing suspicion. INN-based image hiding has become an important research branch, which formulates the hiding and revealing processes as a pair of inverse transformations. However, inherent architectural constraints confine existing INN-based methods to single-scale, single-domain feature modeling and limited nonlinear representation, resulting in inferior image quality. The CrosInv we introduce in this paper addresses these limitations. Compared with the second-best results, CrosInv improves hiding PSNR by 18.086, 18.111, and 18.338 dB on COCO, ImageNet, and BOSSBase, and improves revealing PSNR by 11.443, 11.593, and 13.416 dB. The proposed CrosInv provides a more reliable alternative to existing image hiding solutions.
\end{IEEEImpStatement}

\begin{IEEEkeywords}
Covert communication, image hiding, invertible neural network, spatial-frequency collaboration, non-invertible mechanism.
\end{IEEEkeywords}

\section{Introduction}

\IEEEPARstart{D}{igital} images are widely shared on the Internet, making them common cover media for steganography. Image steganography is the art and science of embedding secret messages into publicly available cover images without arousing suspicion for covert communication \cite{Shi:24,Li:26}. A well-designed steganographic scheme typically offers high capacity, good image quality, and strong resistance to steganalysis. With the development of deep learning, steganographic models have demonstrated remarkable capability, successfully hiding a full-size image in another one, also known as image hiding \cite{Baluja:17}. The bit-level steganographic methods with a relatively small payload often perfectly decode the secret message while remaining undetectable to steganalysis. In contrast, image hiding relaxes the requirement of perfect reconstruction and instead emphasizes the high capacity with a trade-off in image quality between stego images and revealed secret images. Owing to the substantially increased payload, the requirement for resistance to steganalysis is comparatively less stringent.

\begin{figure}[t]
	\centering
	\includegraphics[width=1.0\columnwidth]{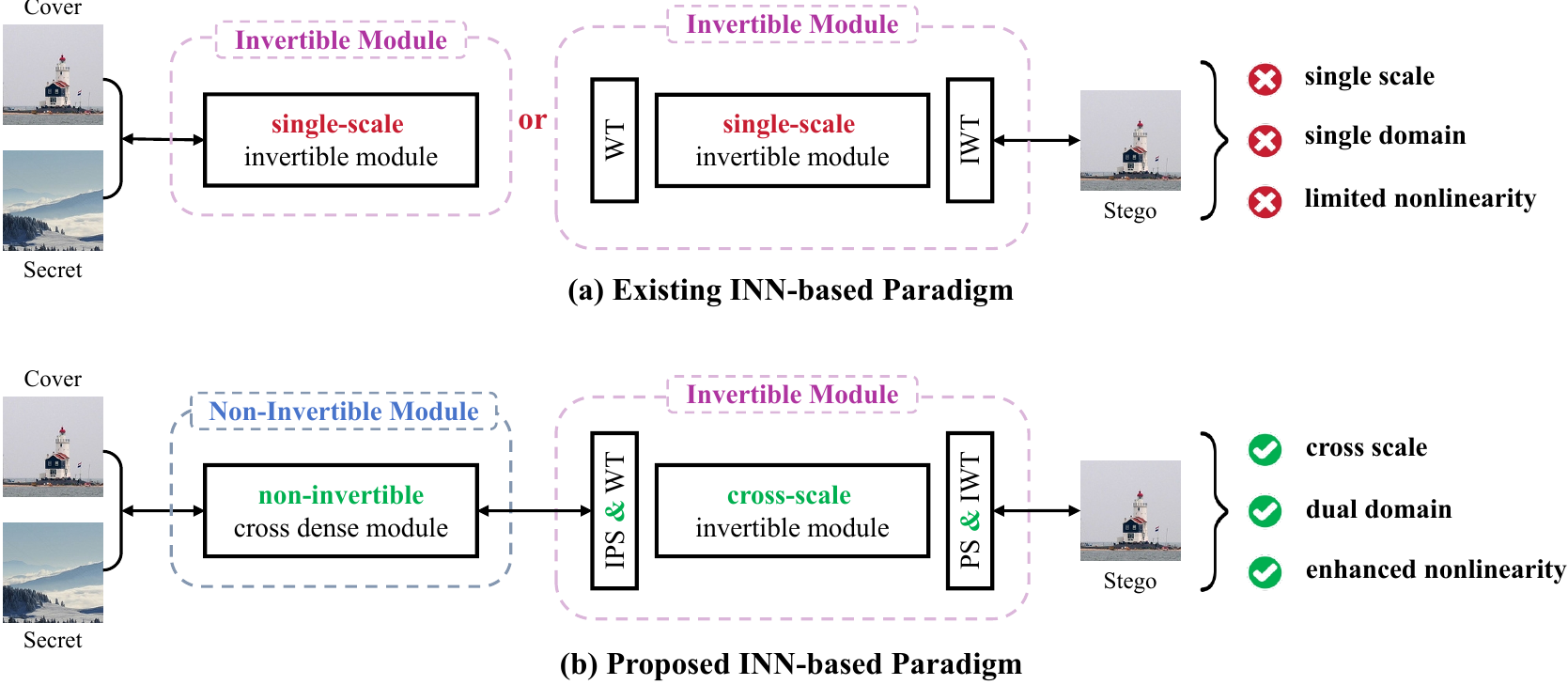} 
	\caption{The design distinctions between our CrosInv and existing INN-based image hiding methods. (a) Existing INN-based methods suffer from single-scale and single-domain feature extraction and limited nonlinear representation capability. (b) The proposed CrosInv exploits cross-scale and spatial-frequency collaborative features while enhancing nonlinear representation. PS denotes pixel shuffle. WT denotes Haar wavelet transformation. IPS and IWT are the inverse operations. PS/IPS and WT/IWT perform scale transformation while introducing spatial and frequency information, respectively.}
	\label{fig1}
\end{figure}

Current image hiding frameworks can be broadly divided into two categories: autoencoder-based and INN-based architectures. Autoencoder-based methods \cite{Baluja:17,Baluja:20,Zhang:20,Zhang:23,Ke:24} typically employ two separate networks for concealing and revealing, respectively. INN-based approaches \cite{Lu:21, Jing:21, Hu:24,Xiong:25, Shi:26} treat concealing and revealing as a pair of inverse problems on image domain transformation and use INN's forward and backward processes to address them. Benefiting from the theoretically lossless property \cite{Dinh:17}, INN-based approaches can better preserve the information during transformation, making them particularly suitable for the image hiding task.

While the INN excels in image hiding, it still faces several limitations. Due to its strict invertibility, INN has limited flexibility and nonlinearity. Specifically, existing INN-based methods \cite{Lu:21, Jing:21, Hu:24,Xiong:25,Shi:26} employ the single-scale invertible module, without exploiting features across different scales. Moreover, current models are implemented in a single domain, typically either the spatial \cite{Lu:21, Hu:24} or wavelet \cite{Jing:21, Xiong:25,Shi:26}, neglecting the collaboration between two domains. Additionally, their limited nonlinearity makes them difficult to fully capture complex feature relationships for image hiding. The overall framework design of these methods is illustrated in Fig.\ref{fig1} (a).

To mitigate the aforementioned limitations, we propose CrosInv, an efficient cross-scale invertible hiding network with the spatial-frequency collaboration and the non-invertible mechanism, as shown in Fig.\ref{fig1} (b). The main idea of CrosInv is to improve INN-based image hiding by jointly exploiting cross-scale and spatial-frequency collaborative features while enhancing nonlinear representation. Unlike conventional INN-based frameworks restricted to a single scale, the proposed method introduces a cross-scale invertible module (CIM) that bijectively maps inputs into cross-scale representations. Such a design is also used for image colorization \cite{Zhao:21} and image enhancement \cite{Quan:24} tasks. However, to the best of our knowledge, it has not yet been explored in the image hiding task. Furthermore, CIM employs pixel shuffle (PS) \cite{Shi:16}, Haar wavelet transformation (WT) \cite{Lienhart:02}, and their inverse operations for scale transformation. PS/IPS provides spatial content, and WT/IWT provides frequency content, naturally introducing spatial and frequency information. Additionally, we propose a non-invertible cross dense module (NCDM), inspired by the dense architecture in \cite{Li:21}. The two NCDMs are symmetrically placed in the forward and inverse processes of CIM to enhance the nonlinearity. Overall, our contributions can be summarized as follows:

\begin{itemize}
	\item We introduce a cross-scale invertible module that bijectively maps inputs into cross-scale representations, enabling cross-scale feature learning beyond the single-scale limitation of the conventional invertible module.
	\item We introduce pixel shuffle, Haar wavelet transformation, and their inverses for scale transformation, during which spatial and frequency information are naturally integrated.
	\item We design the non-invertible cross dense module, symmetrically placed in the forward and inverse processes, respectively, to enhance the nonlinearity.
	\item Comprehensive experiments conducted on three benchmark datasets demonstrate the effectiveness and superiority of the proposed CrosInv.
\end{itemize}

\section{Related Work}

In this section, we summarize advancements in image hiding, including autoencoder-based and INN-based architectures.

\subsection{Autoencoder-Based Image Hiding}

Early studies typically adopt two independent networks to perform the concealing and revealing tasks separately, naming the autoencoder-based architecture. The pioneering method \cite{Baluja:17} proposes a hiding framework composed of three networks: the secret preprocessing component, the secret hiding component, and the secret revealing component. Each component is constructed with vanilla convolutions. The extension method \cite{Baluja:20} further explores the hiding of multiple images, retaining the original network architecture. Building on this framework, UDH \cite{Zhang:20} proposes a cover-agnostic scheme that eliminates the secret preprocessing stage and processes only the secret image during the hiding stage. Its hiding network adopts a simplified U-Net architecture, whereas the revealing network is constructed by stacking vanilla convolutions. DAH-Net \cite{Zhang:23} employs two U-Net-based subnetworks within its hiding network to separately extract the cover and secret features. StegFormer \cite{Ke:24} adopts U-Net-based architectures for both hiding and revealing networks, incorporating Transformer modules to enhance feature representation.

\subsection{INN-Based Image Hiding}

Another category is based on the INN architecture. INN-based approaches treat concealing and revealing as a pair of inverse problems and utilize INN's forward and backward processes to address them. ISN \cite{Lu:21} implements INN in the spatial domain, which consists of a series of affine coupling layers, each incorporating three 5-layer DenseNet blocks. HiNet \cite{Jing:21} integrates INNs with the discrete wavelet transform (DWT) and introduces a low-frequency wavelet loss. Similar to ISN, DIH-OAIN \cite{Hu:24} implements INN in the spatial domain and designs dual-resolution branches within its affine coupling layers to capture image semantics and details. EFCA-DIH \cite{Xiong:25} combines the Sobel operator and DWT to extract edge-aware wavelet subband features, incorporating coordinate attention into affine coupling layers to enhance feature representation. StarINN \cite{Shi:26} also implements INN in the wavelet domain and employs different feature extraction subnetworks for cover and secret images within affine coupling layers to facilitate efficient image hiding and revealing, while introducing a Star operation to enhance feature representation and accelerate model convergence.

Benefiting from the theoretically lossless property \cite{Dinh:17}, INN-based approaches can better preserve information in the image hiding task. However, existing methods typically operate in a single domain, such as the spatial or wavelet domain. Moreover, strict invertibility limits architectural flexibility and nonlinearity. Feature dimensions must maintain consistency across affine coupling layers, while architectural modifications are confined to these layers, restricting feature extraction and representation capability.

\begin{figure*}[t]
	\centering
	{\includegraphics[width=1.0\textwidth]{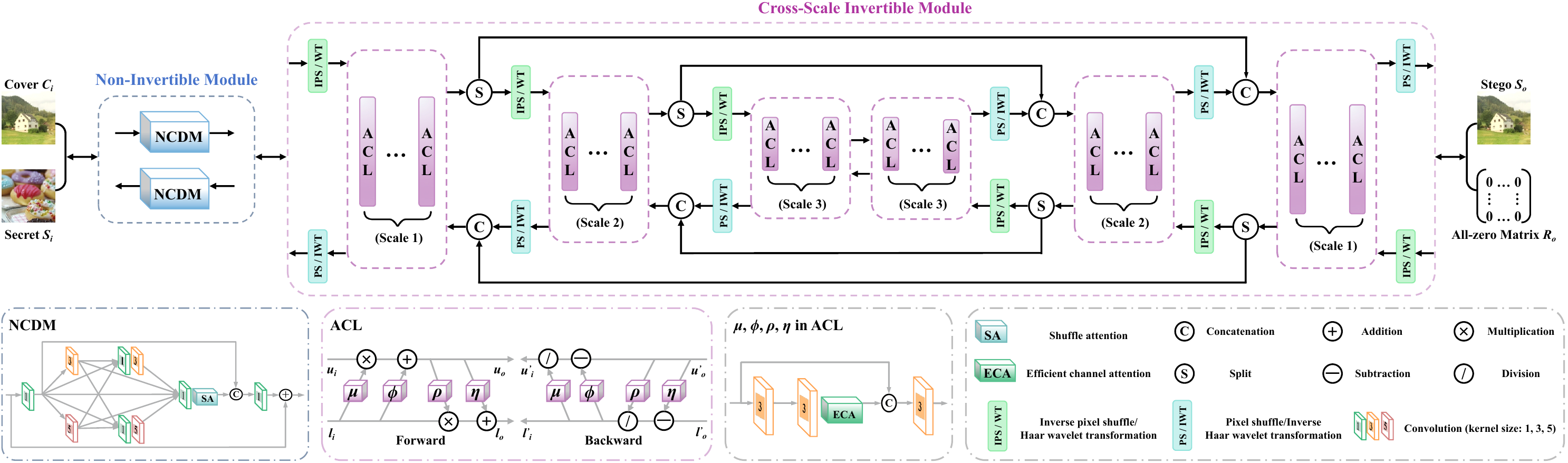}}
	\caption{Overall architecture. CrosInv contains two key components: the non-invertible cross dense module (NCDM) and the cross-scale invertible module (CIM). The two NCDMs are symmetrically placed in the forward and inverse processes of CIM. CIM comprises a series of affine coupling layers (ACLs), the optional scale transformations, and split/concatenation skip connections.}
	\label{fig2}
\end{figure*}

\section{Proposed Framework}

\subsection{Overview}

We propose CrosInv, an efficient cross-scale invertible hiding network that incorporates the spatial-frequency collaboration and the non-invertible mechanism. The main idea of CrosInv is to exploit cross-scale and spatial-frequency collaborative features while enhancing nonlinear representation. The pipeline of CrosInv consists of two key components: the non-invertible cross dense module (NCDM) and the cross-scale invertible module (CIM). The two NCDMs are symmetrically placed in the forward and inverse processes of CIM. The forward process aims to hide the secret image $S_{i}$ within the cover image $C_{i}$ to obtain the stego image $S_{o}$. Conversely, the backward process takes $S_{o}$ and an all-zero matrix $R_{o}$ as input to reveal $S_{i}$. The overall architecture is shown in Fig. \ref{fig2}.

\subsection{CrosInv Model}

{\bf{Non-invertible cross dense module.}} NCDM is utilized to enhance the nonlinear representation capability. As shown in Fig. \ref{fig2}, NCDM starts and ends with a $1\times1$ convolution, enabling efficient channel integration and dimension adjustment. The intermediate part adopts a dual-branch structure that uses two cross dense connection blocks as the backbone, complemented by a $1\times1$ convolution with shuffle attention (SA) \cite{Zhang:21}. The upper branch comprises a $3\times3$ convolution and a pair of $1\times1$ and $3\times3$ convolutions. The lower branch replaces all $3\times3$ convolutions with $5\times5$ ones to enlarge the receptive field and enhance feature representation. The two cross dense connection blocks facilitate the rich feature extraction across different layers and convolutions, and the $1\times1$ convolution with SA adaptively fuses these features.

{\bf{Cross-scale invertible module.}} CIM is exploited to extract cross-scale and spatial-frequency collaborative features. CIM comprises a series of affine coupling layers, the optional scale transformations, and split/concatenation skip connections.

\textbf{\emph{{Affine coupling layer.}}} The affine coupling layer is the basic unit of INN \cite{Dinh:17}. The structure of the adopted affine coupling layer is illustrated in Fig. \ref{fig2}. For the affine coupling layer, the input needs to be divided into two parts, $u_{i}$ and $l_{i}$, along the channel axis. We employ the same enhanced affine transformation for $u_{i}$ and $l_{i}$. Specifically, the forward process of the affine coupling layer can be formulated as
\begin{equation}
	\begin{gathered}	
		u_{o}=u_{i}\odot{\textbf{exp}(\mu(l_{i}))}+\phi(l_{i}) \\
		l_{o}=l_{i}\odot{\textbf{exp}(\rho(u_{o}))}+\eta(u_{o})
		\label{eq1}
	\end{gathered}
\end{equation}
where $\mu(\bigcdot)$, $\phi(\bigcdot)$, $\rho(\bigcdot)$, and $\eta(\bigcdot)$ can be arbitrary structures. Here, we adopt a 3-layer residual block with $3\times3$ convolutions and efficient channel attention (ECA) \cite{Wang:20}, ensuring the representation capability while maintaining lower computational overhead. $\odot$ and $\textbf{exp}(\bigcdot)$ are the element-wise multiplication and the Exponential function, respectively. $\textbf{exp}(\bigcdot)$ is omitted in Fig. \ref{fig2}. Correspondingly, the inverse process is expressed as
\begin{equation}
	\begin{gathered}
		l^{\prime}_{i}=(l^{\prime}_{o}-\eta(u^{\prime}_{o}))\oslash\textbf{exp}(\rho(u^{\prime}_{o})) \\
		u^{\prime}_{i}=(u^{\prime}_{o}-\phi(l^{\prime}_{i}))\oslash\textbf{exp}(\mu(l^{\prime}_{i}))
		\label{eq2}
	\end{gathered}
\end{equation}
where $\oslash$ denotes the element-wise division.

\textbf{\emph{{Scale transformation.}}} CIM employs pixel shuffle (PS) \cite{Shi:16}, Haar wavelet transformation (WT) \cite{Lienhart:02}, and their inverses (IPS and IWT) for scale transformation. These operations are invertible and can facilitate cross-scale representations by rescaling the resolution of the input. Moreover, PS/IPS provides spatial content, and WT/IWT provides frequency content, naturally introducing spatial and frequency information. IPS performs space-to-channel rearrangement with a downscaling factor of 2, and PS performs channel-to-space rearrangement with an upscaling factor of 2. WT performs a single wavelet transformation to downscale (or upscale in IWT) the space dimension by a factor of 2. For example, given a feature map of size $(c, h, w)$, the downscaling operation produces $(4c, h/2, w/2)$, while the corresponding upscaling operation inversely reconstructs the feature map to $(c, h, w)$.

\textbf{\emph{{Split/concatenation skip connections.}}} CIM introduces U-Net-style skip connections at intermediate scales (Scale 2 and Scale 3 in Fig. \ref{fig2}) to capture non-local cross-scale information. Specifically, in the forward/backward process, the previous output is split into two parts. One part is downscaled and passed to the next scale block, and the other is concatenated with the output of the corresponding upscaling operation. The split ratio is 1:1 for IPS and 1:3 for WT.

\subsection{Loss Function}
{\bf{Total loss.}} The total loss $\mathcal{L}_{t}$ consists of two parts: the forward hiding loss $\mathcal{L}_{h}$ to ensure hiding quality, and the backward revealing loss $\mathcal{L}_{r}$ to ensure the revealing quality. $\mathcal{L}_{t}$ is equal to the sum of two losses, expressed as
\begin{equation}
	\mathcal{L}_{t}=\mathcal{L}_{h}+\mathcal{L}_{r}
	\label{eq3}
\end{equation}
where the loss weight is set to 1 by default.

{\bf{Hiding loss.}} The forward hiding process aims to generate a stego image $S_o$ that is close to the cover image $C_i$. Thus, $\mathcal{L}_{h}$ is defined as follows
\begin{equation}	
	\mathcal{L}_{h}=MSE(C_{i}, S_{o})
	\label{eq4}
\end{equation}

{\bf{Revealing loss.}} The backward revealing process aims to losslessly recover the secret image $S_i$ from the generated stego image $S_o$ using an all-zero matrix $R_{o}$. Toward this goal, we define the $\mathcal{L}_{r}$ as
\begin{equation}
	\mathcal{L}_{r}=MSE(S_{i}, S^{\prime}_{i})
	\label{eq5}
\end{equation}
where $S^{\prime}_{i}$ denotes the actual restored secret image. Note that we do not set any restrictions on the all-zero matrix and the recovered cover image, because these two variables are less important in the whole hiding and revealing tasks. This design also encourages CrosInv to concentrate information relevant to secret image recovery in the stego image.

\begin{table*}[]
	\caption{Quantitative comparisons of hiding quality ($C_i,S_o$). CrosInv significantly outperforms existing models. Bold and underline indicate the best and second-best results, respectively.}
	\centering
	\begin{tabular}{ccccccccccccc}
		\toprule
		\multirow{2.5}{*}{Method} &
		\multicolumn{4}{c}{COCO} &
		\multicolumn{4}{c}{ImageNet} &
		\multicolumn{4}{c}{BOSSBase} \\
		\cmidrule(lr){2-5}\cmidrule(lr){6-9}\cmidrule(lr){10-13}
		\multicolumn{1}{c}{}
		& PSNR$\uparrow$ & SSIM$\uparrow$ & APD$\downarrow$ & \multicolumn{1}{c}{LPIPS$\downarrow$} &  PSNR$\uparrow$ & SSIM$\uparrow$ & APD$\downarrow$ & \multicolumn{1}{c}{LPIPS$\downarrow$} &  PSNR$\uparrow$ & SSIM$\uparrow$ & APD$\downarrow$ & LPIPS$\downarrow$ \\ \midrule
		ISN \cite{Lu:21} & 36.768 & .94905 & 2.894 & .00637 & 36.784 & .94938 & 2.932 & .00612 & 38.086 & .94680 & 2.506 & .00909 \\
		HiNet \cite{Jing:21} & 37.505 & .95914 & 2.564 & .00156 & 37.370 & .95665 & 2.693 & .00160 & 38.874 & .95642 & 2.271 & .00212 \\
		StegFormer \cite{Ke:24} & \underline{42.846} & \underline{.98890} & \underline{1.425} & \underline{.00051} & \underline{42.724} & \underline{.98704} & \underline{1.555} & \underline{.00060} & \underline{42.331} & \underline{.97888} & \underline{1.587} & \underline{.00053} \\
		StarINN \cite{Shi:26} & 40.225 & .97998 & 1.957 & .00055 & 40.135 & .97747 & 2.087 & .00063 & 41.010 & .96994 & 1.872 & .00092 \\ 
		CrosInv & \bf{60.760} & \bf{.99940} & \bf{0.104} & \bf{.00004} & \bf{60.649} & \bf{.99930} & \bf{0.123} & \bf{.00006} & \bf{60.517} & \bf{.99936} & \bf{0.103} & \bf{.00007} \\ 
		\bottomrule
	\end{tabular}
	\label{tab:hi}
\end{table*}

\begin{table*}[]
	\caption{Quantitative comparisons of revealing quality ($S_i,S^{\prime}_{i}$). CrosInv is with the optimal performance.}
	\centering
	\renewcommand{\arraystretch}{1}
	\begin{tabular}{ccccccccccccc}
		\toprule
		\multirow{2.5}{*}{Method} &
		\multicolumn{4}{c}{COCO} &
		\multicolumn{4}{c}{ImageNet} &
		\multicolumn{4}{c}{BOSSBase} \\
		\cmidrule(lr){2-5}\cmidrule(lr){6-9}\cmidrule(lr){10-13}
		\multicolumn{1}{c}{}
		& PSNR$\uparrow$ & SSIM$\uparrow$ & APD$\downarrow$ & \multicolumn{1}{c}{LPIPS$\downarrow$} &  PSNR$\uparrow$ & SSIM$\uparrow$ & APD$\downarrow$ & \multicolumn{1}{c}{LPIPS$\downarrow$} &  PSNR$\uparrow$ & SSIM$\uparrow$ & APD$\downarrow$ & LPIPS$\downarrow$ \\ \midrule
		ISN \cite{Lu:21} & 31.505 & .91440 & 5.114 & .05090 & 31.237 & .91001 & 5.436 & .05606 & 32.912 & .90426 & 4.504 & .05734 \\
		HiNet \cite{Jing:21} & 40.107 & .97904 & 1.939 & .00169 & 39.821 & .97615 & 2.115 & .00240 & 39.915 & .96431 & 2.032 & .00217 \\
		StegFormer \cite{Ke:24} & \underline{41.863} & \underline{.98671} & 1.620 & .00072 & \underline{41.610} & \underline{.98432} & \underline{1.783} & \underline{.00083} & 41.164 & .97140 & 1.801 & .00108 \\
		StarINN \cite{Shi:26} & 41.638 & .98564 & \underline{1.609} & \underline{.00069} & 41.316 & .98282 & 1.789 & .00114 & \underline{41.483} & \underline{.97210} & \underline{1.712} & \underline{.00092} \\
		CrosInv & \bf{53.178} & \bf{.99913} & \bf{0.304} & \bf{.00011} & \bf{53.067} & \bf{.99912} & \bf{0.320} & \bf{.00013} & \bf{54.780} & \bf{.99919} & \bf{0.228} & \bf{.00015} \\
		\bottomrule
	\end{tabular}
	\label{tab:re}
\end{table*}

\section{Experiments}
\subsection{Experimental Settings}
{\bf{Datasets.}} In our implementation, we employ three datasets: COCO \cite{Lin:14}, ImageNet \cite{Russakovsky:15}, and BOSSBase \cite{Bas:11}. We use 5000, 1000, and 2000 cover-secret pairs from COCO for training, validation, and testing, respectively. 2000 cover-secret pairs, each from ImageNet and BOSSBase, are used only for testing. For the sake of comparison, the images are center-cropped to the common image sizes of $128\times128$.

{\bf{Baselines.}} We select four representative image hiding methods: ISN \cite{Lu:21}, HiNet \cite{Jing:21}, StegFormer \cite{Ke:24}, and StarINN \cite{Shi:26}. Among them, StegFormer is an autoencoder-based method, while the other three approaches are INN-based. For a fair comparison, we follow ISN and incorporate the same differentiable quantization operation, the straight-through estimator \cite{Bengio:13}, into all methods, converting the generated stego and recovered secret images to 8-bit integers.

{\bf{Evaluation metrics.}} To measure the image quality, we adopt two commonly used objective indicators, peak signal-to-noise ratio (PSNR) and structural similarity index measure (SSIM) \cite{Wang:04}, as well as average pixel discrepancy (APD) calculated by $L1$ norm, and learned perceptual image patch similarity (LPIPS) \cite{Zhanga:18}, which is more consistent with human visual perception. Note that we evaluate the average PSNR value of three channels in the RGB color space. Higher PSNR/SSIM and lower APD/LPIPS values indicate higher image quality. We use three color image steganalyzers, including the hand-crafted features-based SCRMQ1 \cite{Goljan:14} and the deep learning-based UCNet \cite{Wei:22} and PENet \cite{Wei:24}, to evaluate the anti-steganalysis ability. The detection accuracy rate is closer to 50\%, i.e., a random guess, indicating superior security.

{\bf{Implementation details.}} CrosInv is implemented with PyTorch, and the NVIDIA GeForce RTX 4090 GPU is used for acceleration. We adopt the Adam optimizer with $betas=(0.9, 0.999)$ and $eps=1e^{-8}$. The initial learning rate is $1e^{-4}$. After 20 epochs, it is decayed by a factor of 0.5 every 10 epochs. The mini-batch size is 4, containing two randomly selected cover/secret pairs. We adopt three types of scale blocks to construct CrosInv. Given an input image size of 128, the corresponding scales are set to 64 (Scale 1), 32 (Scale 2), and 16 (Scale 3), respectively. Accordingly, we use three downscaling and three upscaling scale blocks. For maintaining lower computational overhead, the number of affine coupling layer in each scale block is only set to 1. Notably, CrosInv converges within 50 epochs and achieves stable performance.

\subsection{Comparison Results}

To comprehensively evaluate the proposed CrosInv, we conduct experiments in terms of visual quality, quantitative metrics, anti-steganalysis ability, and computational overhead.

\begin{figure}[t]
	\centering
	{\includegraphics[width=1.0\columnwidth]{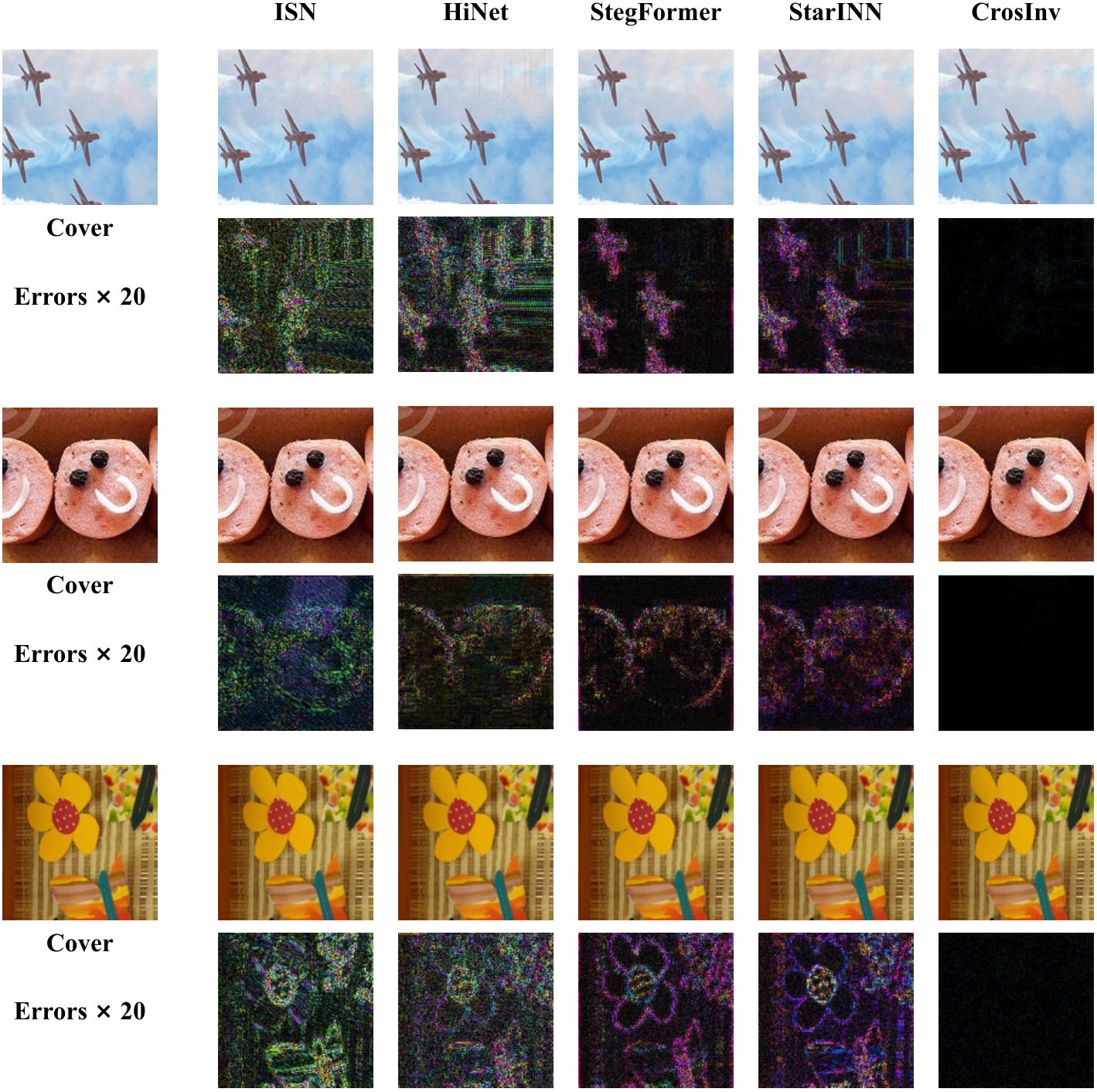}}
	\caption{Visual comparisons of hiding quality ($C_i,S_o$). Stego images generated by CrosInv have the minimal residuals.}
	\label{HI}
\end{figure}

\begin{figure}[t]
	\centering
	{\includegraphics[width=1.0\columnwidth]{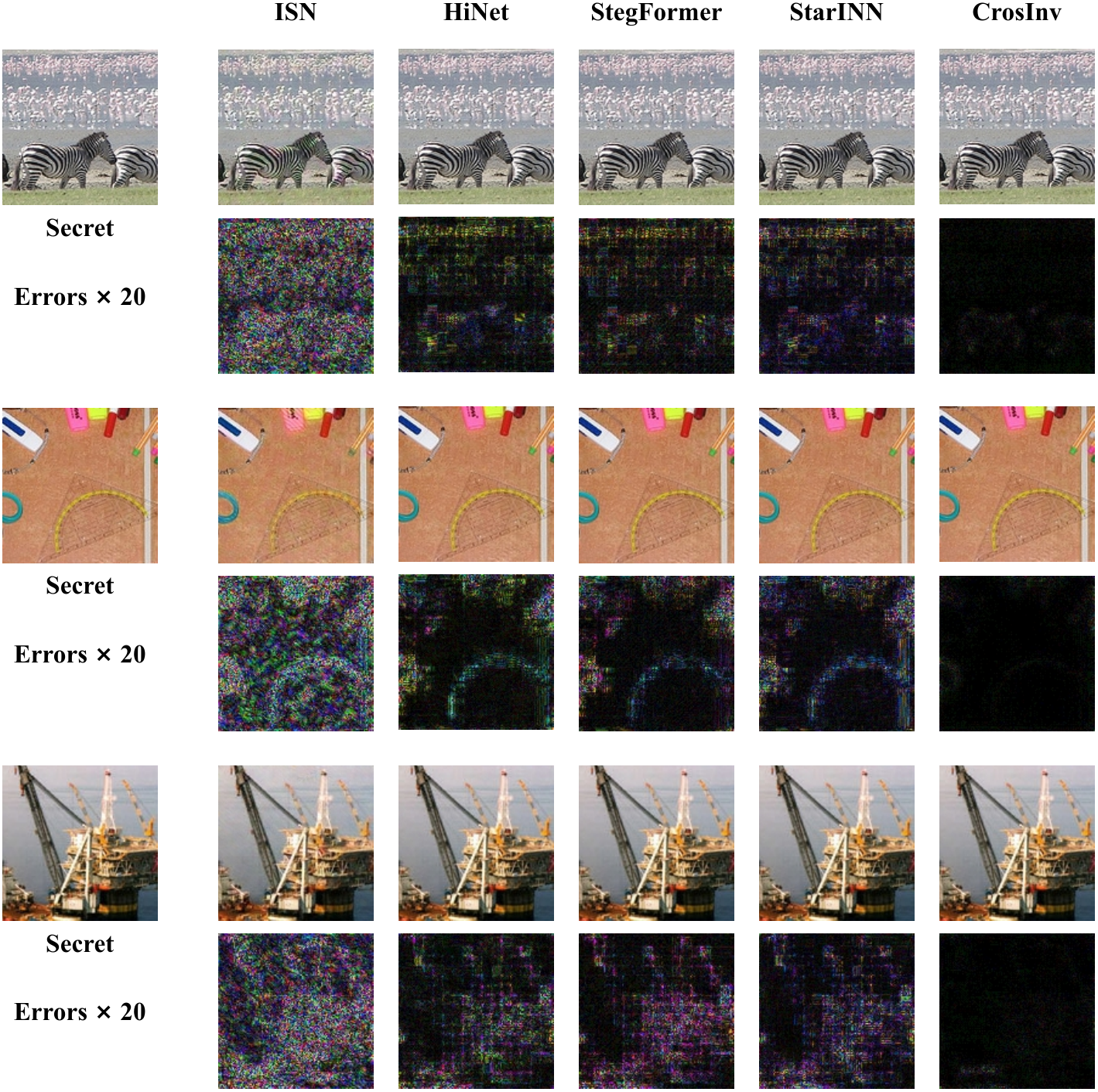}}
	\caption{Visual comparisons of revealing quality ($S_i,S^{\prime}_{i}$). Secret images revealed by CrosInv have the minimal residuals.}
	\label{RE}
\end{figure}

\textbf{Visual quality.} Visualization comparisons between our CrosInv and four representative methods are presented in Figs. \ref{HI} and \ref{RE}. We show three examples of hiding and revealing results, respectively. It can be observed that although these methods produce visually pleasing stego and recovered secret images, our method yields the minimal residuals. The residuals remain at a very low magnitude even when they are amplified by a factor of 20. In contrast, the comparison methods produce more noticeable residuals in both stego and recovered secret images. These results demonstrate that our CrosInv achieves superior hiding and revealing performance.

\textbf{Quantitative metrics.} Tables \ref{tab:hi} and \ref{tab:re} report the quantitative comparisons of our CrosInv with ISN \cite{Lu:21}, HiNet \cite{Jing:21}, StegFormer \cite{Ke:24}, and StarINN \cite{Shi:26} in terms of the four image quality metrics. We can observe that the proposed CrosInv significantly outperforms other approaches in terms of both hiding and revealing performance. For hiding quality, compared with the second-best results, CrosInv achieves PSNR improvements of 18.086 dB, 18.111 dB, and 18.338 dB on COCO, ImageNet, and BOSSBase datasets, respectively. Regarding revealing quality, CrosInv provides PSNR gains of 11.443 dB, 11.593 dB, and 13.416 dB. Similar performance advancements are also seen in SSIM, APD, and LPIPS metrics. Notably, although our model is trained solely on the COCO, it still provides nice results on the other two datasets, demonstrating a favorable cross-dataset generalization ability.

\textbf{Anti-steganalysis ability.} Steganalysis provides a more rigorous criterion by measuring the probability of distinguishing stego images from cover images. Here, we employ a hand-crafted features-based color image steganalyzer, SCRMQ1 \cite{Goljan:14}, and two advanced deep learning-based color image steganalyzers, UCNet \cite{Wei:22} and PENet \cite{Wei:24}. The comparisons on the anti-steganalysis ability are reported in Table \ref{table:UCNet}. We can see that the stego images generated by CrosInv have higher anti-steganalysis ability.

\begin{table}[t]
	\centering
	\caption{Comparisons of anti-steganalysis ability. CrosInv has the best anti-steganalysis ability.}
	\label{table:UCNet}
	\begin{tabular}{cccc}
		\toprule
		\multirow{2.5}{*}{Method} & \multicolumn{3}{c}{Accuracy(\%) $\downarrow$} \\
		\cmidrule(lr){2-4}
		\multicolumn{1}{c}{} & SCRMQ1 \cite{Goljan:14} & UCNet \cite{Wei:22} & PENet \cite{Wei:24} \\ \midrule
		ISN \cite{Lu:21} & \underline{92.385} & \underline{78.995} & \underline{89.660} \\
		HiNet \cite{Jing:21} & 93.585 & 95.665 & 97.418 \\
		StegFormer \cite{Ke:24} & 95.010 & 86.300 & 92.304 \\
		StarINN \cite{Shi:26} & 94.962 & 95.393 & 97.640 \\
		CrosInv & \bf{81.660} & \bf{74.180} & \bf{67.940} \\
		\bottomrule
	\end{tabular}
\end{table}

\begin{table}[t]
	\caption{Comparisons of computational overhead. Params ($\times10^4$), FLOPs ($\times10^8$), Times ($s$). CrosInv can achieve state-of-the-art performance with relatively lower overhead.}
	\centering
	\setlength{\tabcolsep}{4.8mm}
	\begin{tabularx}{\columnwidth}{cccc}
		\toprule
		Method & Params & FLOPs & Times \\
		\midrule
		ISN \cite{Lu:21} & \bf{149.808} & 489.861 & \bf{0.00812} \\
		HiNet \cite{Jing:21} & 810.048 & 662.490 & 0.02650 \\
		StegFormer \cite{Ke:24} & 3492.081 & \bf{96.296} & 0.01833 \\
		StarINN \cite{Shi:26} & 1605.978 & 1308.560 & 0.14877 \\
		CrosInv & \underline{239.969} & \underline{112.764} & \underline{0.01182} \\
		\bottomrule
	\end{tabularx}
	\label{tab:EF}
\end{table}

{\bf{Computational overhead.}} We further discuss the computational overhead by calculating the number of floating point operations (FLOPs), model parameters (Params), and the inference time for a single image (Times). The results in Table \ref{tab:EF} show that the computational overhead of CrosInv is second-best. CrosInv can achieve state-of-the-art performance with relatively lower overhead.

\begin{table}[t]
	\caption{Ablation of cross scale, PS/IPS, WT/IWT, and NCDM. CrosInv benefits notably from the designs of cross scale and NCDM. The collaboration between PS/IPS and WT/IWT can further improve performance.}
	\centering
	\setlength{\tabcolsep}{1.2mm}
	\begin{tabular}{cccccc}
		\toprule
		Cross scale & PS/IPS & WT/IWT & NCDM & PSNR($C_i,S_o$) & PSNR($S_i,S^{\prime}_{i}$) \\
		\midrule
		\ding{55} & \ding{51} & \ding{51} & \ding{51} & 51.357 & 50.145 \\
		\ding{51} & \ding{55} & \ding{51} & \ding{51} & 57.624 & 51.930 \\
		\ding{51} & \ding{51} & \ding{55} & \ding{51} & \underline{59.129} & \underline{52.396} \\
		\ding{51} & \ding{51} & \ding{51} & \ding{55} & 49.388 & 48.497 \\
		\ding{51} & \ding{51} & \ding{51} & \ding{51} & \bf{60.760} & \bf{53.178} \\
		\bottomrule
	\end{tabular}
	\label{tab:ab}
\end{table}

\subsection{Ablation Study}

In this subsection, we investigate the effectiveness of several key design components in CrosInv. Specifically, we consider the effect of cross scale, PS/IPS, WT/IWT, and NCDM. The ablation results are presented in Table \ref{tab:ab}. (1) ``Cross scale (\ding{55})'': All scale transformations in CrosInv are removed. For a fair comparison, the number of blocks is adjusted so that the model capacity remains comparable to that of the original CrosInv. (2) ``PS/IPS (\ding{55})'': All PS/IPS operations are replaced with WT/IWT, i.e., only frequency information is introduced. (3) ``WT/IWT (\ding{55})'': All WT/IWT operations are replaced with PS/IPS, i.e., only spatial information is introduced. (4) ``NCDM (\ding{55})'': The non-invertible module is removed. As we can see from the table, the designs of cross scale and NCDM significantly enhance the learning capability of CrosInv. The collaboration between PS/IPS and WT/IWT can further improve the performance of CrosInv.

\section{Conclusion}

In this work, we present an efficient image hiding model, termed CrosInv, that can exploit cross-scale and spatial-frequency collaborative features while enhancing nonlinear representation. CrosInv comprises two key components: NCDM and CIM. The two NCDMs are symmetrically placed in the forward and inverse processes of CIM to enhance the nonlinear representation capability. CIM comprises a series of affine coupling layers, the optional scale transformations, and split/concatenation skip connections, enabling effective extraction of cross-scale and spatial-frequency collaborative features. Comprehensive experiments verify the effectiveness and superiority of the proposed CrosInv.

\end{document}